%% file: main.tex
\makeatletter
\def\@fnsymbol#1{\ensuremath{\ifcase#1\or \dagger\or \ddagger\or
   \mathsection\or \mathparagraph\or \|\or **\or \dagger\dagger
   \or \ddagger\ddagger \else\@ctrerr\fi}}
\documentclass[10pt,twocolumn,letterpaper]{article}

\usepackage[pagenumbers]{cvpr} % To force page numbers, e.g. for an arXiv version

\input{preamble}

\definecolor{cvprblue}{rgb}{0.21,0.49,0.74}
\usepackage[pagebackref,breaklinks,colorlinks,citecolor=cvprblue]{hyperref}

\def\paperID{6470} % *** Enter the Paper ID here
\def\confName{CVPR}
\def\confYear{2024}

\title{Curriculum Point Prompting for Weakly-Supervised Referring Image Segmentation}

\author{
Qiyuan Dai \quad 
Sibei Yang\thanks{Corresponding author}\\ 
 School of Information Science and Technology, ShanghaiTech University \\
{\tt\small \{daiqy2022,yangsb\}@shanghaitech.edu.cn}\\
}

\begin{document}
\maketitle
\input{camera_ready/sec/0_abstract}    
\input{camera_ready/sec/1_intro}
\input{camera_ready/sec/2_related_V2}

\input{camera_ready/sec/3_method_V2}
\input{camera_ready/sec/5_experiments_V2}
\input{camera_ready/sec/6_conclusion}

{
    \small
    \bibliographystyle{ieeenat_fullname}
    \bibliography{main}
}


\end{document}

%% file: preamble.tex
\usepackage[dvipsnames]{xcolor}

\usepackage{multirow}
\usepackage[accsupp]{axessibility}

%% file: camera_ready/sec/0_abstract.tex
\begin{abstract}

Referring image segmentation (RIS) aims to precisely segment referents in images through corresponding natural language expressions, yet relying on cost-intensive mask annotations. Weakly supervised RIS thus learns from image-text pairs to pixel-level semantics, which is challenging for segmenting fine-grained masks. 
A natural approach to enhancing segmentation precision is to empower weakly supervised RIS with the image segmentation foundation model SAM. Nevertheless, we observe that simply integrating SAM yields limited benefits and can even lead to performance regression due to the inevitable noise issues and challenges in excessive focus on object parts. 
In this paper, we present an innovative framework, \textbf{P}oint \textbf{P}romp\textbf{T}ing (PPT), incorporated with the proposed multi-source curriculum learning strategy to address these challenges. 
Specifically, the core of PPT is a point generator that not only harnesses CLIP's text-image alignment capability and SAM's powerful mask generation ability but also generates negative point prompts to address the noisy and excessive focus issues inherently and effectively. 
In addition, we introduce a curriculum learning strategy with object-centric images to help PPT gradually learn from simpler yet precise semantic alignment to more complex RIS. 
Experiments demonstrate that our PPT significantly and consistently outperforms prior weakly supervised techniques on mIoU by 11.34\%, 14.14\%, and 6.97\% across RefCOCO, RefCOCO+, and G-Ref, respectively.

\end{abstract}

%% file: camera_ready/sec/1_intro.tex
\section{Introduction}

Referring image segmentation (RIS) entails segmenting the referent referred to by a natural language expression in an image~\cite{hu2016segmentation, ding2021vision, feng2021encoder, yang2022lavt, tang2023contrastive,strudel2022weakly}. This task delves into pixel-level semantic understanding that aligns with free-form texts, as opposed to pixel classification into closed-set categories in typical semantic segmentation~\cite{chen2017deeplab, cheng2022masked, long2015fully, xu2023side}. In recent years, RIS has made significant progress, resulting in improved accuracy in the fully supervised learning setting~\cite{ding2021vision, feng2021encoder, yang2022lavt, tang2023contrastive} and demonstrating great potential in real-world applications, including text-based human-robot interaction~\cite{zhang2023large} and image editing~\cite{brooks2023instructpix2pix, zhang2023adding}. 
However, annotating high-quality pairs of natural language expressions and their corresponding referent masks in images is challenging and time-consuming, limiting the development of RIS which relies on full annotation. 

\begin{figure}[t]
    \vspace{0.1cm}
    \centering
    \includegraphics[width=1\linewidth]{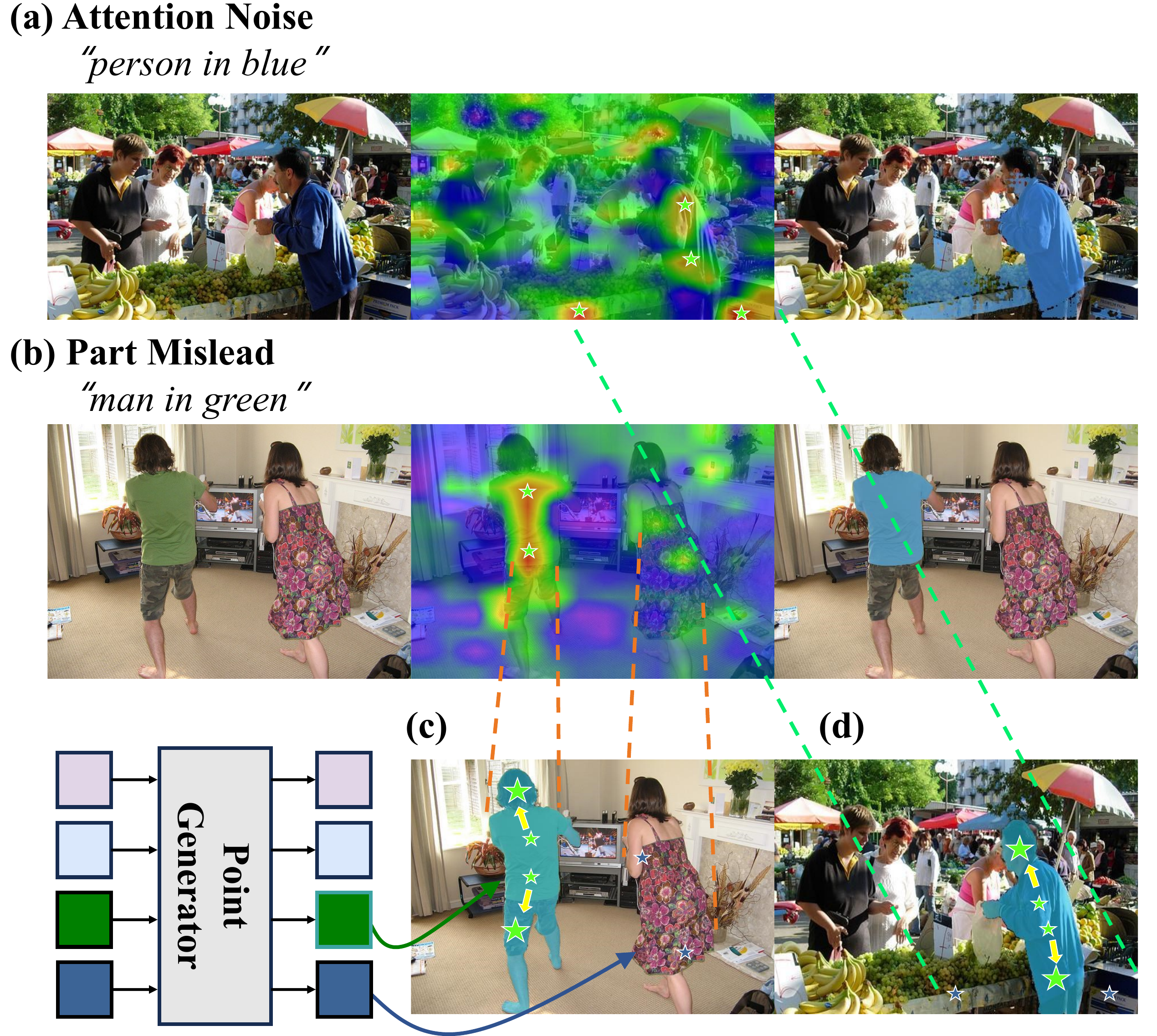}
    \caption{Illustrations of the CLIP attention map on RIS image-text pairs and corresponding mask outputs through SAM. (a) and (b) show background noise issues and excessive focus on object parts that mislead SAM, while (c) and (d) demonstrate the results of our method, which mitigates these issues.}
    \vspace{-6mm}
    \label{fig:intro}
\end{figure}

Therefore, our objective is to study RIS through weakly supervised learning in line with some very recent works~\cite{lee2023weakly, kim2023shatter, strudel2022weakly, liu2023referring}, specifically without the need for any instance-level or pixel-level semantic supervision, thereby mitigating the annotation limitation. 
In light of the absence of pixel-level semantic annotations, the primary focus of these weakly supervised RIS studies is to transfer the semantic associations from image-text pairs to the pixel level. This is achieved by visual entity discovery and gathering~\cite{kim2023shatter}, text-to-image response mapping~\cite{liu2023referring}, or enhancing Grad-CAM for an improved saliency map~\cite{lee2023weakly}. 
While it is possible to approximate the regions of referents, the segmentation results lack precision.

Lately, the Segment Anything Model (SAM)~\cite{kirillov2023segment} has demonstrated its proficiency in generating valid image segmentation masks. A naive approach to improving weakly supervised RIS involves employing SAM to refine the coarse localization of referents into precise segmentation masks. Despite relying on annotated masks, SAM's training does not include pixel-level semantic supervision. This supervision aligns with our weakly supervised RIS setting and offers a promising ``free lunch" solution as an image segmentation foundation model. 
\textcolor{black}{However, unexpectedly, this refinement fails to enhance, or even drop, the performance of RIS.} We observe that the primary reasons for this are: 1) The pixel-level semantic response obtained from image-text pairs inherently contains noise or activates other objects or attributes mentioned in the expression, yet SAM is not robust when using such noisy responses as input prompts directly. As shown in Figure~\ref{fig:intro}(a), the attention response for the expression in the image indeed localizes the referent. Nevertheless, the presence of certain noisy attention, like that in Figure~\ref{fig:intro}(a), results in SAM predicting a segmentation that encompasses all these responsive regions. 2) More crucially, attending salient rather than comprehensive responses in weakly supervised RIS worsens the problem of segmentation ambiguity in SAM, where input prompts can correspond to multiple valid masks. This is due to the combined effect of SAM's almost edge-oriented, semantic-unaware nature and the image-level semantic supervision inherent to weakly supervised RIS. For example, the green t-shirt in Figure~\ref{fig:intro}(b) naturally elicits a more noticeable response for the expression ``man in green". However, using the response as the mask, box, or points to prompt SAM leads to segmenting the t-shirt rather than the man.

In this paper, we propose a novel point prompt learning framework that collaborates with an innovative multi-source curriculum learning strategy to tackle these challenges in weakly supervised RIS. Our framework utilizes frozen CLIP as the text and image encoders, frozen SAM as the mask decoder, and a trainable, lightweight point generator to seamlessly bridge the encoders and the decoder. Specifically, the point generator initializes learnable point queries to represent segmentation masks, which interact with image and text features from the encoders, generating point prompts for SAM to achieve precise mask segmentation. This approach harnesses the inherent advantages of CLIP's pre-trained image-text alignment and SAM's segmentation capabilities, effectively simplifying the RIS problem to learning point queries and selecting the positive query for the referent. Notably, point queries naturally handle noisy responses by distinguishing them as negatives, as illustrated in Figure~\ref{fig:intro}(d). 

Moreover, to address the semantic-unaware ambiguity issue, we introduce training on object-centric images like ImageNet~\cite{deng2009imagenet}, encouraging point queries to shift their attention from salient to comprehensive responses. First, ImageNet boasts a large vocabulary with thousands of classes, and its rich semantics can assist point queries in learning semantic-aware point prediction. Additionally, ImageNet predominantly features object-centric images, where the objects are often centrally positioned, making it easier to extract comprehensive class responses using CLIP or unsupervised DINO models~\cite{radford2021learning, caron2021emerging, oquab2023dinov2}. Training on these comprehensive responses can correct the initial point prompts to match those shown in Figure~\ref{fig:intro}(c), resulting in a precise mask of the ``person". To the best of our knowledge, we are the first to leverage object-centric images to improve scene-centric RIS.

Ultimately, in essence, RIS remains a scene-centric task. It entails segmenting referents in complex scenes, considering not only class semantics as seen in object-centric scenes but also their location and relationships with other objects. Therefore, we introduce a novel curriculum learning strategy that progressively transitions from class-based segmentation to complex referring image segmentation, incorporating factors such as location and relationships. Additionally, beyond the complexity of expressions, object-centric images in ImageNet and scene-centric images in RIS have distinct data distributions. We also strive to mitigate the domain gap when jointly training on these datasets by introducing a multi-source training strategy.  

To evaluate the effectiveness of our PPT and learning strategy, we conduct expressions on common benchmarks in RIS, including RefCOCO~\cite{yu2016modeling}, RefCOCO+~\cite{yu2016modeling}, and G-Ref~\cite{mao2016generation}. In summary, our main contributions are: 
\begin{itemize}
\setlength{\itemsep}{0pt}
\setlength{\parsep}{0pt}
\setlength{\parskip}{0pt}
\item We propose a novel, parameter-efficient point prompting framework (PPT) for weakly-supervised RIS. PPT's core component is a trainable, lightweight point generator to identify the referent and naturally distinguish noise and misleading context as negative prompts, thereby enabling effective integration with frozen CLIP and SAM. 
\item To the best of our knowledge, we are the first to effectively utilize object-centric images to facilitate scene-centric RIS to learn precise and comprehensive dense semantic alignment between text and image. 
\item We propose an innovative multi-source curriculum learning strategy to facilitate the gradual learning of the point generator, starting from simpler semantic alignment and progressing to more complex RIS, which meanwhile mitigates the domain mismatch in multi-source training. 
\item Although straightforward, our PPT consistently outperforms state-of-the-art weakly-supervised RIS by a significant margin of 11.34\%, 14.14\%, and 6.97\% in terms of mIoU on the three benchmarks, respectively. 
\end{itemize}

%% file: camera_ready/sec/2_related_V2.tex
\section{Related Work}

\begin{figure*}[th]
\centering
\includegraphics[width=1.00\textwidth]{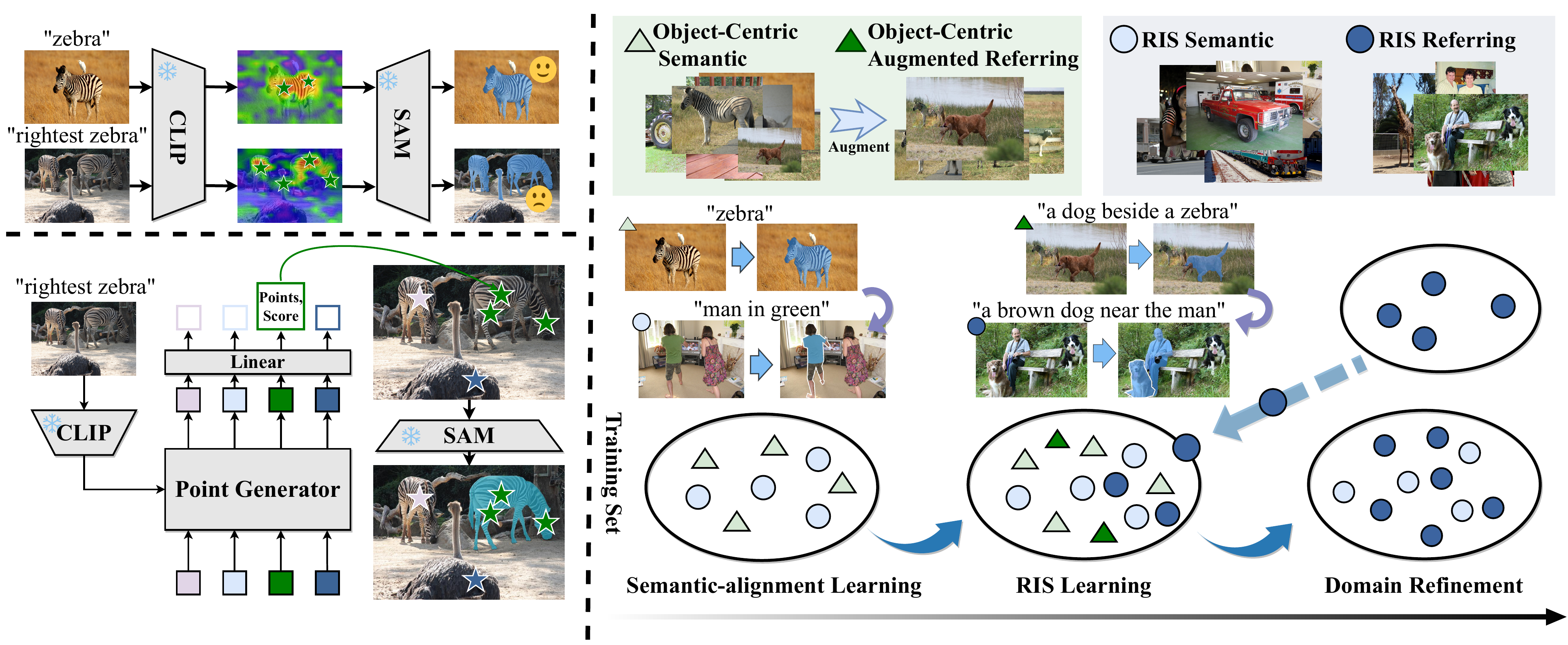}
\vspace{-6mm}
\caption{\textbf{The overall framework of our PPT and its curriculum learning strategy}. The top left corner demonstrates that a straightforward concatenation of CLIP and SAM exhibits segmentation capability for simple object-centric images but performs poorly on referring expression data. The bottom left corner showcases our point generator, which determines the approximate location of objects by learning a well-distributed set of points, combined with our curriculum learning strategy depicted on the right side of the figure, which transitions from simple object-centric semantics towards the more complex referring domain.}
\label{fig:pipeline}
\vspace{-0.2cm}
\end{figure*}

\noindent\textbf{Weakly Supervised RIS}:
RIS aims to segment the precise mask of the referent through a comprehensive understanding of both the image and the text describing the referent, which has achieved significant performance improvements in the fully supervised setting by employing multi-modal fusion from concatenation operation~\cite{hu2016segmentation,li2018referring,liu2017recurrent,shi2018key, yang2019cross, yang2020propagating} to recent attention-based approaches~\cite{kim2022restr, yang2020graph, wang2022cris, yang2022lavt, tang2023contrastive, yang2021bottom, lin2021structured, yang2019dynamic, yang2020relationship}.
However, achieving full supervision demands pixel-level semantic annotations, which entail significant expenses.
In response, recent approaches~\cite{kim2023shatter,liu2023referring, lee2023weakly} begin exploring weakly supervised RIS, leveraging weak supervisory signals like bounding boxes or image-text pairs, aiming to align pixel-level semantics utilizing coarse-grained supervision. 
Among them, Kim {\it et al.}~\cite{kim2023shatter} divides image features into several entities and then employs top-down multi-modal attention to select entities combined into the referring segment. 
Instead of aggregating visual entities, TRIS ~\cite{liu2023referring} extracts rough object positions through text supervision as pseudo-labels to train a segmentation network.
In contrast, Lee {\it et al.}~\cite{lee2023weakly} focuses on reasoning about word relationships to predict from salient maps of each word.
Nevertheless, the masks from these methods are often coarse due to noisy pseudo-labels or lacking dedicated fine segmentation decoding. 
To address this, we introduce a novel framework, PPT, which uses point prompting to handle noise issues, thus enabling the effective utilization of SAM's segmentation capabilities for high-quality masks.

\noindent\textbf{Curriculum Learning for RIS:} 
Curriculum learning proposed by~\cite{bengio2009curriculum}, similar to the natural learning process, gradually increases the complexity of training data during the training to enhance the model's capability. 
Its effectiveness has been validated in computer vision~\cite{shu2019transferable,zhang2021flexmatch,castells2020superloss,yu2020multi} and has recently been introduced into the vision-language field.
For visual reasoning, CLIP-VG~\cite{xiao2023clip} is the first to introduce curriculum learning into visual grounding. It iteratively selects high-quality data for training from a pseudo-label set and uses updated weights for the next round of data selection, achieving a progressive effect.  
For the RIS task, MCRES ~\cite{xu2023meta} combines words from expressions into different novel compositions, enabling the model to learn from word-word, word-phrase, and ultimately phrase-phrase relationships. 
In contrast to previous fully supervised works, we introduce curriculum learning to mitigate the substantial noise present in weakly supervised learning, extending the model's ability from simpler semantic alignment to more complex RIS.

\noindent\textbf{Visual Foundation Models}
possess substantial knowledge capacity, making them versatile for a wide range of tasks. For example, the CLIP model~\cite{radford2021learning} demonstrates significant performance improvements in open-vocabulary tasks~\cite{shi2023edadet, zang2022open,kuo2022f, du2022learning, gu2021open, shi2024the}. In image segmentation, the SAM~\cite{kirillov2023segany} also opens up promising avenues\cite{dai2023samaug, lin2023sequential}. 
In fully supervised RIS, while~\cite{wang2022cris,lai2023lisa, xu2023bridging} generally treat foundation models merely as weight initialization or auxiliary tools for pseudo-label extraction, under-utilizing their intrinsic powerful representation capabilities. Among these, CRIS~\cite{wang2022cris} builds upon CLIP by adding a vision-language decoder to extend into a segmentation model. 
Similarly, LISA~\cite{lai2023lisa} employs SAM's image encoder and LLaVA~\cite{liu2023visual} to train an additional mask decoder. 
Recent, other works like Grounding DINO ~\cite{liu2023grounding} and Grounded SAM~\cite{kirillov2023segany, liu2023grounding} require extensive object detection data during training to provide strong prior knowledge for RIS. However, they contradict the weakly supervised setting. In contrast to prior work, our PPT fully harnesses the capabilities of foundation models in both the encoder and decoder stages and concatenates them through a learnable point generator, simultaneously possessing robust image-text understanding and precise mask generation.

%% file: camera_ready/sec/3_method_V2.tex
\section{Method}

The framework of our proposed PPT and its corresponding multi-source curriculum learning strategy are shown in Figure~\ref{fig:pipeline}. 
First, we introduce PPT, our point prompt learning framework for weakly supervised RIS. Its central focus is on utilizing a trainable, lightweight point generator to transfer the text-image semantic alignment capability from frozen CLIP to SAM, enabling robust and precise mask decoding for referents (see Section~\ref{subsec:framework}). 
Next, we propose learning from object-centric images to support the point generator in generating semantic-aware and comprehensive point prompts, as opposed to merely salient ones (see Section~\ref{subsec:images}). 
Finally, we apply curriculum learning to enable the progressive learning of the point generator, moving from simpler class-based segmentation to more complex referring image segmentation, while also mitigating the domain mismatch in multi-source training (see Section~\ref{subsec:learning}). 

\subsection{Point Prompt Learning Framework}
\label{subsec:framework}

\subsubsection{Point Prompting Architecture}
The overall PPT architecture is simple and illustrated in Figure~\ref{fig:pipeline}. It comprises three primary components: the image and text encoders to extract features, a transformer-based point generator to predict a set of point prompts and their corresponding confidence scores, and a mask decoder that predicts the segmentation from point prompts. 

\noindent\textbf{Image Encoder and Text Encoder.} Follow previous works \cite{subramanian2022reclip, wang2022cris, liu2023referring}, we employ CLIP as both the image and text encoders. 
In contrast to~\cite{liu2023referring} that fully fine-tunes the encoders to adapt weakly supervised RIS, we opt to freeze the encoders. This preserves the pre-trained image-text alignment of CLIP and ensures parameter-efficient tuning for the entire framework. 
Specifically, given a pair of an image $I$ and a natural language expression $G$, we use the encoders to extract visual features $\{V_n\}_{n=1}^N$ at $N$ stages and obtain the contextual feature representation at word level as $T$.

\noindent\textbf{Point Generator.} Inspired by the DETR-based object detection methods~\cite{carion2020end, zhang2022dino, li2022dn, zhu2020deformable}, which represents objects using a set of object queries, we use a set of point queries to represent segmentation masks. These point queries are initially randomly initialized and then interact with image and text features to predict corresponding point prompts, which are subsequently used as input for the mask decoder. 
First, the interactions are performed across multiple encoding stages because image features at different stages may emphasize different levels of semantic information essential for RIS, ranging from simple shapes to complex semantics. At each stage, such as the $n$-th stage, the visual features $V_n$ and text features $T$ are initially fused to update their representations to $V_n^\prime$ and $T_n$ following~\cite{xu2023bridging}. Then, the set of point queries $Q_n$ alternately query the updated image features $V_n^\prime$ and text features $T_n$ by a classical cross-attention layer~\cite{vaswani2017attention} as follows,
\begin{equation}
\begin{aligned}
    Q_{n+1} &= \mathrm{CrossAttn}(\mathrm{CrossAttn}(Q_n, V_n^\prime), T_n). \\
\end{aligned}
\end{equation}
Here, $\mathrm{CrossAttn}$ denotes the cross-attention layer, and the point queries at the first stage are learnable embeddings that are initialized randomly. 

Next, for each point query $q_k \in Q_{N+1}$ at the final stage, we regress its point prompt $P_k=\{(p_k^m, r_k^m)\}_{m=1}^M$, consisting of $M$ points, and predict its referent confidence score $c_k$ as follows,
\begin{equation}
\begin{aligned}
    p_k^m, r_k^m = \mathrm{PointHead}(q_k), c_k=\mathrm{ScoreHead}(q_k), \\
\end{aligned}
\end{equation}
where $\mathrm{PointHead}$ and $\mathrm{ScoreHead}$ are two independent multilayer perceptions for predicting the points and confident scores, respectively. Here, $p_k^m$ denotes the normalized coordinates of the $m$-th point of the $k$-th point query, while $r_k^m$ signifies the score for classifying as a positive point.

\noindent\textbf{Mask Decoder.} We utilize SAM as the mask decoder and freeze it, similar to our approach with CLIP, to retain its segmentation capability. The mask decoder can be prompted with a set of points, which can be positive or negative, for SAM to infer the area to be segmented and output the corresponding segmentation mask. Specifically, for the $k$-th point query, we employ the point prompt $P_k$ to prompt the mask decoder, leading to the generation of the segmentation mask $s_k$ within image $I$. 

To summarize, given an input image $I$ and an expression $G$, our PPT framework outputs a fixed-size of $K$ predictions $\{y_k\}_{k=1}^K$, where the $k$-th query's prediction $y_k=(s_k, P_k, c_k)$ includes the segmentation mask $s_k$, point prompt $P_k=\{(p_k^m, r_k^m)\}_{m=1}^M$, and confidence score $c_k$, which is formulated as follows,
\begin{equation}
\begin{aligned}
  \{y_k\}_{k=1}^K = f_{\theta}(I, G), \; \text{where} \; y_k=(s_k, P_k, c_k)
\end{aligned}
\end{equation}
Here, we denote our framework as $f_{\theta}$, where the $\theta$ represents learnable parameters in the point generator.  

\subsubsection{Objective Function}
\label{subsubsec:loss}
Inspired by DETR in object detection, we apply the Hungarian algorithm~\cite{kuhn1955hungarian} to search for the best bipartite matching between the pseudo labels $\{\hat{y_{e}}\}_{e=1}^E$ and the predictions $\{y_{k}\}_{k=1}^K$ to determine the best assignments $\sigma(e)$, where $\sigma(e)$ represents the index of the prediction matched with the pseudo label $\hat{y_e}$. The extraction for pseudo labels from image-text pairs will be discussed in Section~\ref{subsec:images} and \ref{subsec:learning}. Specifically, for each pair of the pseudo label $\hat{y}_e=(\hat{s_e}, \hat{P_e}, \hat{c_e})$ and the prediction $y_{\sigma(e)}= (s_{\sigma(e)}, P_{\sigma(e)}, c_{\sigma(e)})$, we define the objective function by considering the segmentation loss related to the point prompt and the mask prediction, as well as the loss for confidence score as the referent. The loss $\ell(y_{\sigma(e)}; \hat{y}_e)$ for the prediction $y_{\sigma(e)}$ given the pseudo label $\hat{y}_e$ is formulated as follows,
\begin{equation}
\label{eq:loss}
\begin{aligned}
\ell_{\mathrm{seg}}( y_{\sigma(e)}; \hat{y}_e) &= \ell_{\mathrm{mask}}(s_{\sigma(e)}; \hat{s}_e) +  \ell_{\mathrm{pt}}(P_{\sigma(e)}; \hat{P_e}),  \\
\ell_(y_{\sigma(e)}; \hat{y}_e) &= \ell_{\mathrm{seg}}( y_{\sigma(e)}; \hat{y}_e) + \ell_{\mathrm{bce}}(c_{\sigma(e)}; \hat{c}_e),
\end{aligned}
\end{equation}
where $\ell_{\mathrm{mask}}$ represents the DICE loss~\cite{li2019dice} and binary cross-entropy loss. The point prompt loss $\ell_{\mathrm{pt}}$ is defined as the combination of $L1$ loss for point regression and the binary cross-entropy loss for point classification. Additionally, $\ell_{\mathrm{bce}}$ denotes the binary cross-entropy loss for classifying the prediction as the referent or not. 

During inference, we select the segmentation mask $s_{\mathrm{argmax}\{c_e\}}$ corresponding to the highest confidence score
as the segmentation prediction for the referent.

\subsection{Learning from Object-centric Images}
\label{subsec:images}

To train our PPT model using image-text pairs, we generate pseudo-labels for referents from these pairs leveraging the pre-trained image-text alignment of CLIP~\cite{radford2021learning}. 
However, directly extracting pseudo-labels from image-expression pairs in RIS datasets~\cite{yu2016modeling, mao2016generation} for training does not yield satisfactory results. The main reasons for this are as follows:  
1) The pixel-level semantic responses obtained through image-text alignment are inevitably noisy or incomplete (particularly in salient regions)
, as shown in Figure~\ref{fig:intro}(a)-(b).
2) The complexity of referring expressions adds to the challenge of extracting their responses as effective pseudo-labels, primarily because distinguishing the response corresponding to the referent from other responses becomes arduous. Referring expressions describe rich content, including class, attributes, locations, and relationships. The semantic response to such expressions encompasses information about the referent but also incorporates additional contextual details, such as the ``a brown dog near the man" in Figure~\ref{fig:pipeline}.

We propose to address these challenges by training on the ImageNet~\cite{deng2009imagenet}, as its object-centric characteristics make it particularly suitable for extracting high-quality pseudo-labels. The top left of Figure~\ref{fig:pipeline} shows that the pseudo segmentation mask for the object-centric image is precise but noisy for the RIS image.
In detail, we construct pseudo datasets for weakly supervised RIS at both the class-semantic level and the more complex relationship level.

\noindent\textbf{Free Data for Simpler Semantic-alignment Learning.}  
We create a simpler semantic alignment dataset, denoted as $D_{\mathrm{sem}}$, from ImageNet for weakly supervised RIS. The dataset is designed to mitigate issues related to noise and partial segmentation by including comprehensive pseudo-masks corresponding to object classes. 

Given an image and its class in ImageNet, we first create an image-expression pair $(I, G)$ by randomly applying transformations, such as flipping, to the image and defining the expression as ``the [class name] in the middle". 
Then, we automatically obtain the corresponding pseudo-label $\{\hat{y}_e\}_{e=1}^E$, where $\hat{y}_e=(\hat{s}_e, \hat{P}_e, \hat{c}_e)$, and $\hat{s}_e$ and $\hat{P}_e$ denote the pseudo segmentation mask and point prompt, respectively. Note that for semantic-level data, we only obtain the referent's pseudo label, which means $E=1$ for the dataset $D_{\mathrm{sem}}$, and score $\hat{c}_e=1$. 
For the segmentation mask $\hat{s}_e$, we employ the CLIP\cite{li2023clip} followed by SAM to directly obtain a high-quality precision mask, benefiting from the object-centric characteristics of the image. 
For the point prompt $\hat{P}_e$, we randomly and uniformly sample $M$ points within the bounding box of the mask $\hat{s}_e$. Points inside the mask are considered positive, while those outside are considered negative. Building upon this sampling approach, we encourage the model to produce points that are evenly distributed within referents and emphasize the negative points closely outside the boundary of referents, significantly reducing the ambiguity in SAM's point prompting.

\noindent\textbf{Augmented Data for More Complex RIS Learning.} 
Training exclusively on semantic-level data could result in a limited understanding of the more complex relationships needed for RIS. As ImageNet itself does not inherently contain any information about relationships between instances, we propose augmenting the semantic-level dataset $D_{\mathrm{sem}}$ to create a referring-level dataset $D_{\mathrm{ref}}$. In dataset $D_{\mathrm{ref}}$, we focus solely on spatial relationships between instances, with the generalization to semantic relationships discussed in Section~\ref{subsec:learning}. 
Specifically, we mainly employ two types of relation augmentation as follows: 
1) In composition-based augmentation, where multiple images are aggregated into one by scaling and pasing through Mosaic-like~\cite{bochkovskiy2020yolov4} augmentation, resulting in crowded images containing multiple instances, along with text that describes absolute positions for the target referent.  
2) In fusion-based augmentation, we embed one image within another and create a text for the target instance that encodes their relative relationship.

Through relation augmentation, our dataset $D_{\mathrm{ref}}$ not only enriches the instance-level semantic data by introducing more complex image-expression pairs, but it also provides pseudo-labels for context objects other than the referent. These contextual pseudo-labels are crucial for learning to distinguish the referent from other objects mentioned in the expressions. For a sample $(I, G, \{\hat{y}_e\}_{e=1}^E) \in D_{\mathrm{ref}}$, its pseudo-labels $\{\hat{y}_e\}_{e=1}^E$ are related with $E$ contextual objects, with each $\hat{y}_e=(\hat{s}_e, \hat{P}_e, \hat{c}_e)$. In these pseudo-labels, the confidence score is set to one for the referent and zero for other context objects.

\subsection{Multi-source Curriculum Learning Strategy}
\label{subsec:learning}

In this section, we present a multi-source curriculum learning strategy for jointly training our PPT model $f_{\theta}$ using both multi-source (ImageNet-based and RIS-based) and multi-level (semantic-level and referring-level) datasets. We first introduce the pseudo-label extraction in the RIS dataset. Then, we present the progressively learning stages from the semantic level to the referring level to domain refinement. At each stage, we utilize different sets of data samples from ImageNet-based datasets $D_\mathrm{sem}$ and $D_\mathrm{ref}$, as well as the RIS dataset $H$.

\noindent\textbf{Pseudo Label Generation.} 
As discussed in Section~\ref{subsec:images}, due to the complexity of referring expressions, extracting pseudo-label for the referent directly from the entire expression results in poor label quality. Therefore, for an image-expression pair $(I, G)$, we extract its pseudo-label $\{\hat{y}_e\}_{e=1}^E$ by collecting all $E$ candidate referents. This is achieved by extracting candidate pseudo-label separately for each noun phrase in the expression using the same method employed in extracting pseudo-label in $D_\mathrm{sem}$. 
Subsequently, based on the count of candidate referents in one pair, we partition the RIS dataset $H$ into two subsets, denoted as $H_{\mathrm{sem}}$ and $H_{\mathrm{ref}}$. 
In dataset $H_{\mathrm{sem}}$, a sample $(I, G, \{\hat{y}_e\}_{e=1}^E) \in H_{\mathrm{sem}}$, with each $\hat{y}_e=(\hat{s}_e, \hat{P}_e, \hat{c}_e)$, contains only one candidate referent, which implies $E=1$ for the $H_{\mathrm{sem}}$ dataset and the score $\hat{c}_1=1$ for this referent. 
For each sample $(I, G, \{\hat{y}_e\}_{e=1}^E) \in H_{\mathrm{ref}}$, it contains multiple candidate referents, \ie, $E>1$. At this stage of pseudo-label extraction, we set the confidence scores $\hat{c}_e=0$ for all candidate referents because the referent cannot be distinguished from the other candidate referents. 
Notably, despite the pseudo-labels being generated in datasets $H_{\mathrm{sem}}$ and $H_{\mathrm{ref}}$, they are highly noisy and necessitate collaboration with ImageNet-based datasets $D_{\mathrm{sem}}$ and $D_{\mathrm{ref}}$ for model training.

\noindent\textbf{Semantic-alignment Learning Stage.} 
At this learning stage, our goal is to jointly train the model using multi-source semantic-level datasets, including $H_\mathrm{sem}$ and $D_\mathrm{sem}$, to predict semantic-aware and comprehensive segmentation masks for references. The inclusion of the dataset $D_\mathrm{sem}$ helps mitigate issues related to noise and partial segmentation pseudo-labels in $H_\mathrm{sem}$. The learning objective is formulated as follows,
\begin{equation}
\theta_{\mathrm{sem}} = 
{\operatorname{argmin}}_{\theta}
 \;
\mathbb{E}_{(I, G, \{\hat{y}_e\}) \sim H_\mathrm{sem} \cup D_\mathrm{sem}}[\ell(f_{\theta}(I, G); \{\hat{y}_e\})],
\end{equation}
where the loss $\ell(f_{\theta}(I, G); \{\hat{y}_e\})$ between the model prediction $f_{\theta}(I, G)$ and the pseudo-label $\{\hat{y}_e\}_{e=1}^E$ is defined in Equation~\ref{eq:loss}. 
And we denote the learned parameters after optimization at this stage as $\theta_{\mathrm{sem}}$.

\noindent\textbf{RIS Learning Stage.} Furthermore, we enable progressive learning, transitioning from simpler semantic-level segmentation to more complex referring image segmentation with relation modeling. This is achieved through two consecutive learning steps: 1) Training exclusively on the ImageNet-based referring-level dataset $D_\mathrm{ref}$, with training samples $(I, G, \{\hat{y}_e\})$ sampled from $ H_\mathrm{sem} \cup D_\mathrm{sem} \cup D_\mathrm{ref}$, leads to the learned parameters as $\theta_{\mathrm{refD}}$. 
2) Updating the pseudo-labels in the referring-level RIS dataset $H_\mathrm{ref}$ and conducting joint training on all four datasets. Specifically, we select pseudo-referents from the candidate referents in $H_\mathrm{ref}$ to update it to $H_\mathrm{ref}^\prime$ using the model $f_{\theta}$ with model parameter ${\theta_{\mathrm{refD}}}$. 
Notably, even though the augmented ImageNet-based dataset $D_\mathrm{ref}$ only considers spatial relationships, we observe that the model learned on it can generalize to predict referents in $H_\mathrm{ref}$ with semantic relationships, thanks to the generalization ability of frozen CLIP encoders. The second training step is formulated as follows,
\begin{equation}
\begin{aligned}
&H_{\mathrm{ref}}^\prime  \leftarrow  \mathrm{Select}(H_{\mathrm{ref}}; f_{\theta}(\cdot ; {\theta_{\mathrm{refD}}})), \\
&\theta_{\mathrm{ref}}={\operatorname{argmin}}_{\theta}
 \; 
 \mathbb{E}_{(I, G, \{\hat{y}_e\}) \sim D_{\mathrm{all}}} [\ell(f_{\theta}(I, G; \theta_{\mathrm{refD}}); \{\hat{y}_e\})],
\end{aligned}
\end{equation}
where $\mathrm{Select}$ represents the selection of pseudo-referents from the candidate referents in $H_{\mathrm{ref}}$ according to the model $f_{\theta}$ with model parameter ${\theta_{\mathrm{refD}}}$, the $D_{\text{all}}$ is the union of the datasets $H_\mathrm{sem}$, $D_\mathrm{sem}$, $D_\mathrm{ref}$, and $H^\prime_\mathrm{ref}$. And the $\theta_{\mathrm{refD}}$ in $f_{\theta}(I, G; \theta_{\mathrm{refD}})$ represent the model's parameters are initialized with $\theta_{\mathrm{refD}}$. The parameter after optimization at this stage is denoted as $\theta_{\mathrm{ref}}$.

\noindent\textbf{Domain Refinement Stage.} 
Despite data augmentation, disparities in data distribution and domain persist between the ImageNet-based and RIS-based datasets. Furthermore, the model $f_{\theta_{\mathrm{ref}}}$ has already acquired point generation capabilities at both the instance and relationship levels through the use of ImageNet-based data in previous stages. Hence, we exclusively fine-tune on the RIS datasets at this stage by continuously adjusting the selected pseudo-referents. The optimization objective is formulated as follows,
\begin{equation}
\begin{aligned}
&H_{\mathrm{ref}}^{(i)} \leftarrow  \mathrm{Select}(H_{\mathrm{ref}}^{(i-1)}; f_{\theta}(\cdot; {\theta^{(i-1)}}), \; H^{(i)}=H_{\mathrm{ref}}^{(i)} \cup H_{\mathrm{sem}}, \\
&\theta^{(i)}={\operatorname{argmin}}_{\theta}
 \; 
 \mathbb{E}_{(I, G, \{\hat{y}_e\}) \sim H^{(i)}} [\ell(f_{\theta}(I, G; \theta^{(i-1)}); \{\hat{y}_e\})].
\end{aligned}
\end{equation}
Here, $H^{(i)}$ and $\theta^{(i)}$ represent the dataset and learned model parameters after the $i$-th round of adjustments. At the first round, the referring-level dataset and parameters are $H_{\mathrm{ref}}^\prime$ and $\theta_{\mathrm{ref}}$, obtained from the RIS learning stage.

%% file: camera_ready/sec/5_experiments_V2.tex
\section{Experiments}
\input{camera_ready/tables/table1_V2}
\input{camera_ready/tables/table2}

\subsection{Datasets and Implementation Details}
\noindent\textbf{Datasets.} 
We conduct experiments on three major datasets: RefCOCO\cite{yu2016modeling}, RefCOCO+\cite{yu2016modeling}, and G-Ref~\cite{mao2016generation}. All the images used in these datasets are from subsets of MSCOCO \cite{lin2014microsoft}, and they respectively contain 142,209, 141,564, and 104,560 sentences. RefCOCO+ does not include absolute directional words, while G-Ref contains longer sentences. Following previous works \cite{yang2022lavt,tang2023contrastive}, we divide RefCOCO and RefCOCO+ into the training set, validation set, testA, and testB. As for G-Ref, we use the validation split by Google. 

\noindent\textbf{Implementation Details.}
We employ the CLIP VIT/B-16 as the encoder following~\cite{lee2023weakly, kirillov2023segment} and use the smallest ViT-B SAM for mask decoder. 
We adopt the Adam\cite{kingma2014adam} optimizer with a batch size of 64 and a learning rate of 0.001. The training is conducted for 50 epochs, of which 10, 30, and 10 epochs are for the semantic alignment, RIS learning and domain refinement stages, respectively. 
We randomly select 2,836 images from the Mini Imagenet~\cite{ravi2016optimization} as our auxiliary data. 
We follow previous approaches to use the mIoU (Mean Intersection over Union) and precision at the 0.3, 0.5, and 0.7 thresholds of mIoU as our main evaluation metrics.

\subsection{Comparison with the State-of-the-Art Methods}

Table \ref{tab:1} compares our PPT with other state-of-the-art approaches across all dataset partitions. Our PPT consistently outperforms all other weakly supervised RIS methods, achieving an average mIoU improvement of 11.34\%, 14.14\%, and 6.97\% on RefCOCO, RefCOCO+, and G-Ref, respectively, over the previous highest performance. 

Compared to TRIS~\cite{liu2023referring}, which shares a strategy of first extracting pseudo-labels and then training with us, we achieve significant improvements in mIoU on the three datasets, with gains of 15.59\%, 14.44\%, and 6.97\%, respectively. This indicates that our point prompting framework, which employs a progressive curriculum learning strategy from multi-source data, is more effective in discovering and localizing targets. 
Furthermore, for a fair comparison, we also integrate SAM into open-sourced TRIS. 
However, such integration leads to a decline in performance due to the inevitable noise issues and challenges in excessive focus on object parts in weakly supervised RIS, which will be detailed in the Appendix. 
In contrast, our approach outperforms these methods by 20.18\%, 19.12\%, 16.31\%, respectively, reflecting that our point generator effectively eliminates noise by utilizing negative prompts.
Compared to Shatter~\cite{kim2023shatter}, which requires finding various parts of the instance and aggregating them to prediction, similar to our point-prompting design, we outperform it by 11.34\%, 16.96\%, and 17.1\%, respectively. This demonstrates that by leveraging object-centric images from ImageNet, our method can predict precise semantic alignment regions instead of partial ones.

As shown in Table \ref{tab:tab2}, our PPT significantly improves precision at different IoU thresholds compared to other weakly supervised methods. Especially at prec@0.5 and prec@0.7, we improve state-of-the-art methods by 25.2\% and 27.2\%, respectively, which reflects the accuracy of our point generator in locating referents and the robustness to scattering points. This is achieved by the curriculum learning strategy and augmented object-centric data, contributing to improving fine-grained mask segmentation.

\begin{figure*}[t]
    \vspace{0.1cm}
    \centering
        \vspace{-6mm}
    \includegraphics[width=0.98\linewidth]{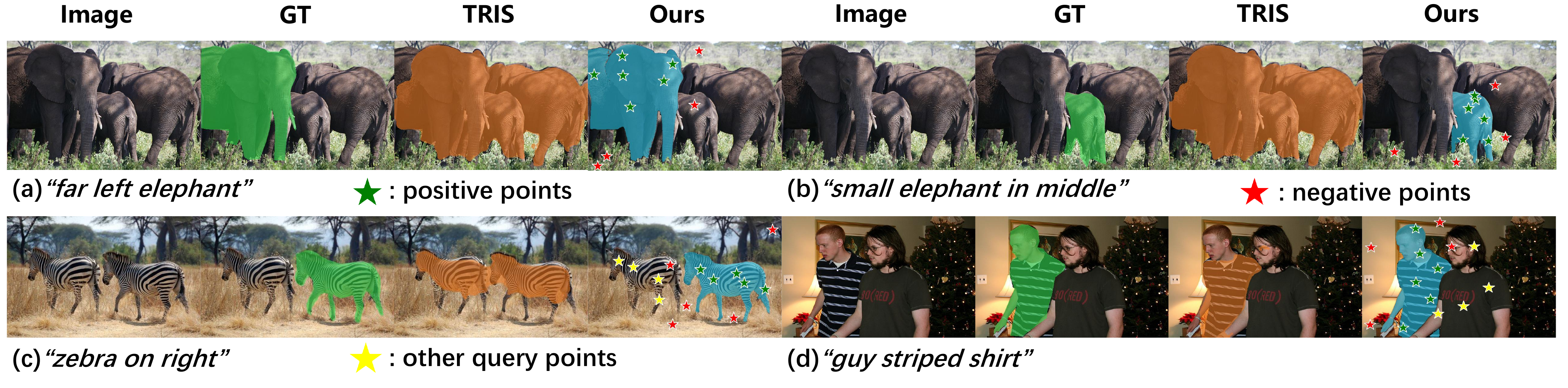}
    \vspace{-4mm}
    \caption{Visualization of our PPT's prediction results and comparison to TRIS~\cite{liu2023referring}.}
    \label{fig:case}
\end{figure*}

\subsection{Ablation Study}

\input{camera_ready/tables/table3}

We conduct ablation experiments to analyze the effectiveness of our proposed method, as shown in Table~\ref{tab:tab3}. 

\noindent\textbf{Point Prompting Framework.} 
(1) We evaluate the performance of directly using CLIP + SAM for RIS in a baseline experiment (row 1). We find that it fails to locate objects in a significant number of images, and even when CLIP identifies the correct region, interference caused by the resolution issue of attention map and background noise suppress SAM mask generation performance. 
(2) Building this foundation, we incorporate the point prompting framework (row 2), achieving a 7.73\% mIoU boost. Our point generator directly utilizes high-dimensional features to regress object positions, thereby circumventing the noise introduced by the low resolution of activation maps. Moreover, it includes negative point prompts, which effectively suppress background interference. 

\noindent\textbf{Curriculum Learning.} 
(3) We perform first-stage training for our prompting framework and gain 4.72\% improvement (row 3), showing the effectiveness of semantic-alignment learning.  
(4) Next, we continue by augmenting training with RIS data (row 4), which aims at enhancing the learning of relation concepts, achieving a 5.23\% mIoU gain. 
The augmented RIS-stage learning helps acquire rudimentary, more complex relation understanding capability. And the curriculum learning strategy filters out noisy data to enhance the training samples' signal-to-noise ratio.
(5) Finally, we introduce the domain refinement stage into our full model, resulting in a further 2.62\% mIoU improvement (row 5), which is achieved by eliminating the domain mismatch during the final fine-tuning phase. 
(6) Row 6 illustrates that without curriculum learning, the mIoU drops by 3.48\%, demonstrating the effectiveness of curriculum learning to select the high-quality pseudo-labels progressively.

\noindent\textbf{Learning from Object-centric Images.}
(7) To assess the impact of object-centric data, we remove augmented data only (row 7), resulting in a 4.47\% mIoU drop, underscoring the assistance our constructed dataset provides in the relation understanding for referring scenarios. 
(8) In row 8, we remove free semantic-alignment data but still keep the augmented data, which resulted in a 7.46\% mIoU drop. This is due to the detrimental impact of incorrect pseudo-label cases on training, which our additional data can mitigate. In addition, it demonstrates that a more straightforward semantic alignment is crucial for follow-up referring segmentation. 
(9) Row 9 demonstrates the results obtained when training without any object-centric images, compared to row 2 and row 5, highlighting that our method achieves maximum gains when both augmented data and curriculum learning are utilized, indicating that curriculum learning can effectively leverage the knowledge from the object-centric images to scene-centric RIS.

\subsection{Visualization}
Figure \ref{fig:case} visualizes our segmentation results. 
Expression in (a) requires the segmentation of the leftmost elephant among the three, demonstrating our framework's ability to distinguish instances. In (b), the expression calls for the segmentation of the middle elephant, showcasing our positional understanding capability. In (c), we visualize the points output by other queries as well, and observe that they also locate the regions of interest within their respective instances. In (d), we show that the point query retrieves object position based on textual semantics, rather than relying solely on spatial pronouns.

%% file: camera_ready/tables/table1_V2.tex
\begin{table*}[t]
   \centering
   \vspace{-4mm}
   \resizebox{2\columnwidth}{!}{
   \setlength{\tabcolsep}{3mm}{\begin{tabular}{l@{}c|c|c|c|c|c|c|c|c|c}
      \toprule[1pt]
      \multirow{2}{*}{Method} & \multirow{2}{*}{Published on} & \multirow{2}{*}{Sup.} & \multirow{2}{*}{Extra Image-Text Pairs} &  
      \multicolumn{3}{c|}{RefCOCO}  & \multicolumn{3}{c|}{RefCOCO+} & \multicolumn{1}{c}{G-Ref} \\
      \cline{5-11}
      &&&& val   & test A & test B & val & test A & test B & val-G  \\
      \hline

      RMI~\cite{liu2017recurrent}  & ICCV '17 & F & - &  44.33 & 44.74 & 44.63 & 29.91 & 30.37 & 29.43 & 33.11 \\
      DMN\cite{margffoy2018dynamic}  & ECCV '18  & F & - &  49.78 & 54.83 & 45.13 & 38.88 & 44.22 & 32.29 & 34.52    \\
      Hu {\it et al.}~\cite{hu2020bi}  & CVPR '20  & F & - &  60.98 & 62.99 & 59.21 & 48.17 & 52.32 & 42.11 & 47.57 \\
      
      GroupViT~\cite{xu2022groupvit}  & CVPR '22  & W & CC12M~\cite{changpinyo2021conceptual},YFCC~\cite{thomee2016yfcc100m} & 12.97 & 14.98 & 12.02 & 13.21 & 15.08 & 12.41 & 16.84 \\

      TSEG~\cite{strudel2022weakly}  & arXiv '22     & W & ImageNet-21K~\cite{deng2009imagenet} & 25.44 & - & - & 22.01 & - & - & 22.05 \\
      ALBEF~\cite{li2021align}  & NeurIPS '21  & W & CC~\cite{sharma2018conceptual},SBU~\cite{ordonez2011im2text},COCO~\cite{lin2014microsoft},VG~\cite{krishna2017visual} & 23.11 & 22.79 & 23.42 & 22.44 & 22.07 & 22.51 & 24.18  \\
      Chunk~\cite{lee2023weakly}  & ICCV '23  & W &  CC,SBU,COCO,VG &  31.06 & 32.30 & 30.11 & 31.28 & 32.11 & 30.13 & 32.88   \\
      Shatter~\cite{kim2023shatter}  & ICCV '23  & W &  ImageNet-21K &  34.76 & 34.58 & 35.01 & 28.48 & 28.60 & 27.98 & 28.87  \\
      TRIS~\cite{liu2023referring}  & ICCV '23  & W & WebImage Text~\cite{radford2021learning} & 31.17 & 32.43 & 29.56 & 30.90 & 30.42 & 30.80 & 36.00  \\
      TRIS + SAM  & -  &  - & WebImage Text &  25.56 &  26.56 & 25.71 & 26.22  & 25.75  & 26.62  & 29.84 \\

      \cline{5-11}
      \rule{0pt}{10pt} 
      \textbf{Our PPT} & CVPR'24 & W & WebImage Text, ImageNet-1K~\cite{russakovsky2015imagenet} &   \textbf{46.76} & \textbf{45.33} & \textbf{46.28} & \textbf{45.34} & \textbf{45.84} & \textbf{44.77} & \textbf{42.97} \\ 
      \midrule[1pt]
\end{tabular}}}
   \caption{Comparison with state-of-the-art models in weakly supervised RIS on RefCOCO, RefCOCO+ and G-Ref datasets.}
   \label{tab:1}
\end{table*}

%% file: camera_ready/tables/table2.tex
\begin{table}[]
\centering
\small
\renewcommand{\arraystretch}{1.07}
\tabcolsep=0.143cm
\begin{tabular}{l|c|ccc|c}
\toprule[1pt]
\multicolumn{1}{l}{Method} &\multicolumn{1}{|c}{Params} &\multicolumn{1}{|c}{P@0.3} &\multicolumn{1}{c}{P@0.5}& \multicolumn{1}{c|}{P@0.7} &\multicolumn{1}{c}{mIoU} \\
\hline
TRIS    &142.09M& 46.57    &   18.69  & 4.9       & 31.17 \\
Shatter  &-& 55.02    & 24.99    & 6.35       & 34.76 \\
Chunk  &-& 46.12    & 23.88    & 9.02       & 31.06 \\
Ours &20.7M& 61.16    & 50.19    & 36.22       & 46.76\\

\bottomrule[1pt]
\end{tabular}
\caption{Comparison of precision metrics. Params denote the number of trainable parameters. }
\label{tab:tab2}
\vspace{-0.1cm}
\end{table}

%% file: camera_ready/tables/table3.tex
\begin{table}[]
\centering
\small
\renewcommand{\arraystretch}{1.07}
\tabcolsep=0.143cm
\begin{tabular}{l|l|ccc|c}
\toprule[1pt]
\multicolumn{1}{c|}{} &Method &\multicolumn{1}{c}{P@0.3} &\multicolumn{1}{c}{P@0.5}& \multicolumn{1}{c|}{P@0.7} &\multicolumn{1}{c}{mIoU} \\
\hline
1&CLIP + SAM                                  & 28.72    & 24.84    & 11.67       & 26.46       \\ 
2&Pt Prompt   w/o IMG                   &   43.85  &  25.45   &    18.35    &  34.19   \\
3&Semantic Learning                      &   51.69  &  39.63   &   28.44     &  38.91  \\ 
4&3+RIS Learning                         &  59.42  &  48.53   &   35.62    &  44.14 \\
5&4+Domain Refine (Full)                              &  61.16  &  50.20   &    36.22   &   46.76  \\
\hline
6&5    w/o CL           & 57.93 & 44.64 &   34.27  & 43.28  \\
7&5    w/o Augmented IMG                & 56.56&  45.19 &   33.03  &  42.29 \\ 
8&5    w/o Free IMG                &  47.29   &  38.51   &    27.94   & 39.33 \\ 
9&5    w/o IMG                    &  43.78 & 27.11   &   18.77   & 34.47 \\

\bottomrule[1pt]
\end{tabular}
\caption{Ablation study on the validation set of RefCOCO. IMG represents Object-centric Images from the ImageNet dataset.}
\label{tab:tab3}
\vspace{-2mm}
\end{table}

%% file: camera_ready/sec/6_conclusion.tex
\section{Conclusion}
We introduce a novel point prompting framework (PPT) for weakly supervised referring image segmentation. It leverages a point generator to connect frozen CLIP and SAM and incorporates the concept of curriculum learning to this field. The framework utilizes object-centric images to aid in learning dense semantic alignment and relationships between text and images in a weakly supervised setting, which helps naturally mitigate noise issues and excessive focus on object parts. 

\noindent\textbf{Acknowledgment:} This work was supported by the National Natural Science Foundation of China (No.62206174) 
and MoE Key Laboratory of Intelligent Perception and Human-Machine Collaboration (ShanghaiTech).